\documentclass[10pt,twocolumn,letterpaper]{article}

\usepackage{cvpr}
\usepackage{times}
\usepackage{epsfig}
\usepackage{graphicx}
\usepackage{amsmath}
\usepackage{amssymb}
\usepackage{subfigure}

\usepackage{authblk}

\usepackage[pagebackref=true,breaklinks=true,letterpaper=true,colorlinks,bookmarks=false]{hyperref}

\cvprfinalcopy % *** Uncomment this line for the final submission

\def\cvprPaperID{2305} % *** Enter the CVPR Paper ID here

\ifcvprfinal\fi

\usepackage[inline]{enumitem}

\usepackage{array}
\newcolumntype{L}[1]{>{\raggedright\let\newline\\\arraybackslash\hspace{0pt}}m{#1}}
\newcolumntype{C}[1]{>{\centering\let\newline\\\arraybackslash\hspace{0pt}}m{#1}}
\newcolumntype{R}[1]{>{\raggedleft\let\newline\\\arraybackslash\hspace{0pt}}m{#1}}

\begin{document}
\title{Deep Multimodal Representation Learning from Temporal Data}

% \author{Xitong Yang\thanks{Work carried out while at PARC, a Xerox Company}\\
% University of Maryland\\
% {\tt\small xyang35@cs.umd.edu}
% \and
% Palghat Ramesh\\
% PARC\\
% {\tt\small Palghat.Ramesh@parc.com}
% \and
% Radha Chitta\footnotemark[1],  \space Sriganesh Madhvanath\footnotemark[1]\\
% Conduent Labs US\\
% {\tt\small \{Radha.Chitta, Sriganesh.Madhvanath\}@conduent.com}
% \and
% Edgar A. Bernal\footnotemark[1]\\
% United Technologies Research Center\\
% {\tt\small bernalea@utrc.utc.com}
% \and
% Jiebo Luo\\
% University of Rochester\\
% {\tt\small jluo@cs.rochester.edu}
% }

\makeatletter
\renewcommand\AB@affilsepx{ \space\space\space \protect\Affilfont}
\author[1]{Xitong Yang\thanks{Work carried out while at PARC, a Xerox Company}}
\author[2]{Palghat Ramesh}
\author[3]{Radha Chitta\protect\footnotemark[1]}
\author[3]{Sriganesh Madhvanath\protect\footnotemark[1]}
\author[4]{\protect\\ Edgar A. Bernal\protect\footnotemark[1]}
\author[5]{Jiebo Luo}
\affil[1]{University of Maryland, College Park}
\affil[2]{PARC}
\affil[3]{Conduent Labs US\protect\\}
\affil[4]{United Technologies Research Center}
\affil[5]{University of Rochester\protect\\}

 \renewcommand\AB@affilsepx{, \protect\Affilfont}
 \affil[1]{\tt\small xyang35@cs.umd.edu}
 \affil[2]{\tt\small Palghat.Ramesh@parc.com}
 \affil[3]{\tt\small \{Radha.Chitta, Sriganesh.Madhvanath\}@conduent.com}
 \affil[4]{\tt\small bernalea@utrc.utc.com}
 \affil[5]{\tt\small jluo@cs.rochester.edu}

\renewcommand\Authands{ and }

\maketitle
\begin{abstract}
In recent years, Deep Learning has been successfully applied to multimodal learning problems, with the aim of learning useful joint representations in data fusion applications. When the available modalities consist of time series data such as video, audio and sensor signals, it becomes imperative to consider their temporal structure during the fusion process. In this paper, we propose the \textit{\textbf{Corr}elational \textbf{R}ecurrent \textbf{N}eural \textbf{N}etwork} (CorrRNN), a novel temporal fusion model for fusing multiple input modalities that are inherently temporal in nature. Key features of our proposed model include: (i) simultaneous learning of the joint representation and temporal dependencies between modalities, (ii) use of multiple loss terms in the objective function, including a maximum correlation loss term to enhance learning of cross-modal information, and (iii) the use of an attention model to dynamically adjust the contribution of different input modalities to the joint representation. We validate our model via experimentation on two different tasks: video- and sensor-based activity classification, and audio-visual speech recognition. We empirically analyze the contributions of different components of the proposed CorrRNN model, and demonstrate its robustness, effectiveness and state-of-the-art performance on multiple datasets.
% * <radha.cr@gmail.com> 2016-11-07T16:38:13.041Z:
%
% Refer to the abstract in the Experiments google doc (https://docs.google.com/document/d/1_YqjPZD_rN57U1pgUULUn-DJC9LKdargL0jZm9xG4xo/edit?usp=drive_web). In this paper, I dont think we are addressing synchronous and asynchronous data.  Also, the experiments are on activity classification and audio-visual speech recognition.
%
% ^ <srig@acm.org> 2016-11-12T03:22:25.147Z.
\end{abstract}

% Xitong
% The figure need to be updated
%
\begin{figure}
\begin{center}
\includegraphics[width=0.8\columnwidth]{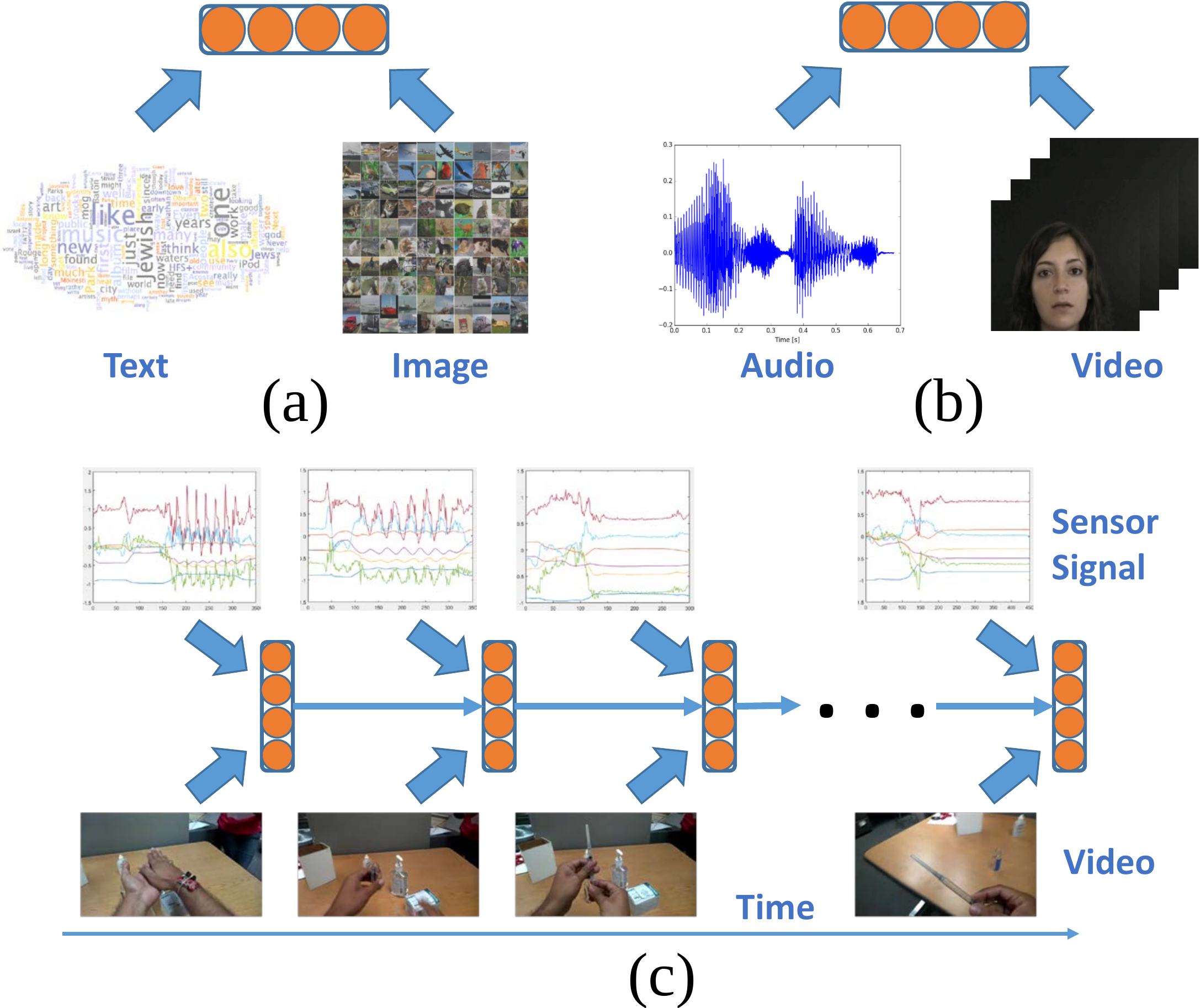}
\end{center}
\caption{Different multimodal learning tasks. (a) Non-temporal model for non-temporal data \cite{srivastava2012multimodal}. (b) Non-temporal model for temporal data \cite{ngiam2011multimodal}. (c) Proposed CorrRNN model: temporal model for temporal data.}
\label{fig:fig1}
\vspace{-1em}
\end{figure}

\section{Introduction}
\label{sec:intro}

%% Paragraph 1: Why temporal multimodal learning
Automated decision-making in a wide range of real-world scenarios often involves acquisition and analysis of data from multiple sources. For instance, human activity may be more robustly monitored using a combination of video cameras and wearable motion sensors than with either sensing modality by itself. When analyzing spontaneous socio-emotional behaviors, researchers can use multimodal cues from video, audio and physiological sensors such as electro-cardiograms (ECG) \cite{ringeval2015av+}. However, fusing information from different modalities is usually nontrivial due to the distinct statistical properties and highly non-linear relationships between low-level features \cite{srivastava2012multimodal} of the modalities. Prior work has shown that \textit{multimodal learning} often provides better performance on tasks such as retrieval, classification and description \cite{kiros2014multimodal,ngiam2011multimodal, srivastava2012multimodal,neverova2016moddrop}. When the modalities being fused are temporal in nature, it becomes desirable to design a model for \textit{temporal multimodal learning} (TML) that can simultaneously fuse the information from different sources, and capture temporal structure within the data.

%% Paragraph 2: Current solutions and why they are unsatisfactory
In the past five years, several deep learning based approaches have been proposed for TML, in particular, for audio-visual data. Early models proposed for audiovisual speech recognition (AVSR) were based on the use of non-temporal models such as deep multimodal \textit{autoencoders} \cite{ngiam2011multimodal} or deep \textit{Restricted Boltzmann Machines} (RBM) \cite{srivastava2012multimodal,sui2015listening} applied to concatenated data across a number of consecutive frames. More recent models have attempted to model the inherently sequential nature of temporal data, \textit{e.g.}, Conditional RBMs \cite{amer2014multimodal}, Recurrent Temporal Multimodal RBMs (RTMRBM) \cite{hu2016temporal} for AVSR, and Multimodal Long-Short-Term Memory networks for speaker identification \cite{ren2016look}.

We believe that a good model for TML should simultaneously learn a joint representation of the multimodal input, and the temporal structure within the data. Moreover, the model should be able to dynamically weigh different input modalities to enable emphasis on the more useful signal(s) and to provide robustness to noise, a known weakness of AVSR \cite{katsaggelos2015audiovisual}. Third, the model should be able to generalize to different kinds of multimodal temporal data, not just audio-visual data. Finally, the model should be tractable and efficient to train. In this paper, we introduce the \textit{\textbf{Corr}elational \textbf{R}ecurrent \textbf{N}eural \textbf{N}etwork} (CorrRNN), a novel unsupervised model that satisfies the above desiderata.

%% Paragraph 3: Our approach

An interesting characteristic of multimodal temporal data from many application scenarios is that the differences across modalities stem largely from the use of different sensors such as video cameras, motion sensors and audio recorders, to capture the same temporal phenomenon. In other words, modalities in multimodal temporal data are often different representations of the same phenomena, which is usually not the case with other multimodal data such as images and text, which are related because of their shared high-level semantics. Motivated by this observation, our CorrRNN attempts to explicitly capture the \textit{correlation} between modalities through maximizing a correlation-based loss function, as well as minimizing a reconstruction-based loss for retaining information. 

This observation regarding correlated inputs has motivated previous work in multi-view representation learning using the Deep Canonically Correlated Autoencoder (DCCAE) \cite{wang2015deep} and Correlational Neural Network \cite{chandar2015correlational}. Our model extends this work in two important ways. First, an RNN-based encoder-decoder framework that uses Gated Recurrent Units (GRU) \cite{ChoMGBBSB14} is introduced to capture the  temporal structure, as well as long-term dependencies and correlation across modalities. Second, dynamic weighting is used while encoding input sequences to assign different weights to input modes based on their contribution to the fused representation. %The dynamic weighting mechanism also controls the choice of correlation for the simple intuition that we should not maximize the correlation of two modalities if their noise level differ too much. %The implementation of the dynamic weighting mechanism is inspired by the attention models proposed in recent works \cite{BahdanauCB14} and can be interpreted as enabling the model to attend on the one with more useful signal if the other one is corrupted with noise. 

The main contributions of this paper are as follows: 
\begin{itemize}
\item We propose a novel generic model for temporal multimodal learning that combines an Encoder-Decoder RNN framework with Multimodal GRUs, a multi-aspect learning objective,  and a dynamic weighting mechanism;
\item We show empirically that our model outperforms state-of-the-art methods on two different application tasks: video- and sensor-based activity classification and audio-visual speech recognition; and  
\item Our proposed approach is more tractable and efficient to train compared with RTMRBM and other probabilistic models designed for TML.
\end{itemize}

The remainder of this paper is organized as follows. In Sec. \ref{sec:related}, we review the related work on multimodal learning. We describe the proposed CorrRNN model in Sec. \ref{sec:model}. Sec. \ref{sec:data} introduces the two application tasks and datasets used in our experiments. In Secs. \ref{sec:exp1} and \ref{sec:exp2}, we present empirical results demonstrating the robustness and effectiveness of the proposed model. The final section presents conclusions and future research directions.

\section{Related work}
\label{sec:related}
In this section, we briefly review some related work on deep-learning-based multimodal learning and temporal data fusion. Generally speaking, and from the standpoint of dynamicity, fusion frameworks can be classified based on the type of data they support (\textit{e.g.}, temporal vs. non-temporal data) and the type of model used to fuse the data (\textit{e.g.}, temporal vs. non-temporal model) as illustrated in Fig. \ref{fig:fig1}.

\subsection{Multimodal Deep Learning}
Within the context of data fusion applications, deep learning methods have been shown to be able to bridge the gap between different modalities and produce useful joint representations \cite{ngiam2011multimodal,srivastava2012multimodal}. Generally speaking, two main approaches have been used for deep-learning-based multimodal fusion. The first approach is based on common representation learning, which learns a joint representation from the input modalities. The second approach is based on \textit{Canonical Correlation Analysis} (CCA) \cite{hardoon2004canonical}, which learns separate representations for the input modalities while maximizing their correlation.

An example of the first approach, the Multimodal Deep Autoencoder (MDAE) model \cite{ngiam2011multimodal}, is capable of learning a joint representation that is predictive of either input modality. This is achieved by performing simultaneous self-reconstruction (within a modality) and cross-reconstruction (across modalities). Srivastava \textit{et al.} \cite{srivastava2012multimodal} propose to learn a joint density model over the space of multimodal inputs using \textit{Multimodal Deep Boltzmann Machines} (MDBM). Once trained, it is able to infer a missing modality through Gibbs sampling and obtain a joint representation even in the absence of some modalities. This model has been used to build a practical AVSR system \cite{sui2015listening}. Sohn \textit{et al.} \cite{sohn2014improved} propose a new learning objective to improve multimodal learning, and explicitly train their model to reason about missing modalities by minimizing the variation of information.

CCA-based methods, on the other hand, aim to learn separate features for the different modalities such that the correlation between them is mutually maximized. They are commonly used in multi-view learning tasks. In order to improve the flexibility of CCA, Deep CCA (DCCA) \cite{andrew2013deep} was proposed to learn nonlinear projections using deep networks. Weirang \textit{et al.} \cite{wang2015deep} extended this work by combining DCCA with the multimodal deep autoencoder learning objective \cite{ngiam2011multimodal}. The Correlational Neural Network model \cite{chandar2015correlational} is similar in that it integrates two types of learning objectives into a single model to learn a common representation. However, instead of optimizing the objective function under the hard CCA constraints, it only maximizes the empirical correlation of the learned projections.

% Our temporal multimodal learning models are extensions of previous works, with the specific application in fusing temporal modalities. In addition to the advantages of temporal modeling, our models optimize an Approximated CCA loss function, which is more scalable to large dataset than classical CCA methods, but retaining the benefits of the original CCA constraints.

%%
%Learning a common representation for multiple modalities is a natural way to integrate data from different sources, and the learnt common representation can be used in any applications, as shown in [7]. However, the aforementioned models suffer from some drawbacks. First, the bimodal deep autoencoder model [7] fails to yield better performance than the unimodal one in the cross modality learning task, which may due to the inability to share the capacity of the common hidden layer between the modalities [10]. Second, the probabilistic models proposed in [8, 9] is more difficult to train since their objective functions are intractable [11]. One problem of these models is that CCA method is not easily scalable to very large dataset. To that end, stochastic optimization methods [15] are proposed for training a deep CCA model.
%%

\subsection{Temporal Models for Multimodal Learning}

In contrast to multimodal learning using non-temporal models, there is little literature on fusing temporal data using temporal models. Amer \textit{et al.} \cite{amer2014multimodal} proposed a hybrid model for fusing audio-visual data in which a \textit{Conditional Restricted Boltzmann Machines} (CRBM) is used to model short-term multimodal phenomena and a discriminative \textit{Conditional Random Field} (CRF) is used to enhance the model. In more recent work \cite{hu2016temporal}, the \textit{Recurrent Temporal Multimodal RBM} was proposed which learns joint representations and temporal structures. The model yields state-of-the-art performance on the ASVR datasets AVLetters and AVLetters2. A supervised multimodal LSTM was proposed in \cite{ren2016look} for speaker identification using face and audio sequences. The method was shown to be robust to both distractors and image degradation by modeling long-term dependencies over multimodal high-level features.

\section{Proposed Model}
\label{sec:model}
In this section, we describe the proposed CorrRNN model. We start by formulating the temporal multimodal learning problem mathematically. For simplicity, and without loss of generality, we consider the problem of fusing two modalities $X$ and $Y$; it should be noted, however, that the model seamlessly extends to more than two modalities. We then present an overview of the model architecture, which consists of two components: the multimodal encoder and the multimodal decoder. We describe the multimodal encoder, which extracts the joint data representation, in Sec. \ref{sec:encoder}, and the multimodal decoder, which attempts to reconstruct the individual modalities from the joint representation in Sec.~\ref{sec:decoder}.

\subsection{Temporal Multimodal Learning}
Let us denote the two temporal modalities as sequences of length $T$, namely $\mathcal{X}=(x_1^{m},x_2^{m},...,x_T^{m})$ and $\mathcal{Y}=(y_1^{n},y_2^{n},...,y_T^{n})$, where $x_t^{m}$ denotes the $m$ dimensional feature of modality $X$ at time $t$. For simplicity, we omit the superscripts $m$ and $n$ in most of the following discussion.

In order to achieve temporal multimodal learning, we fuse the two modalities at time $t$ by considering both their current state and history. Specifically, at time $t$ we append the recent per-modality history to the current samples $x_t$ and $y_t$ to obtain extended representations $\tilde{x_t}=\left \{x_{t-l},...,x_{t-1},x_{t} \right\}$ and $\tilde{y_t}=\left \{y_{t-l},...,y_{t-1},y_{t} \right\}$, where $l$ denotes the scope of the history taken into account. Given pairs of multimodal data sequences $\left\{ \left( \tilde{x_i}, \tilde{y_i} \right) \right\}_{i=1}^N$, our goal is to train a feature learning model $\mathcal{M}$ that learns a $d$-dimensional joint representation $\left\{ \tilde{h_i} \right\}_{i=1}^N$ which simultaneously fuses information from both modalities and captures underlying temporal structures.

\subsection{Model Overview}
\begin{figure}
\begin{center}
\includegraphics[width=.85\columnwidth]{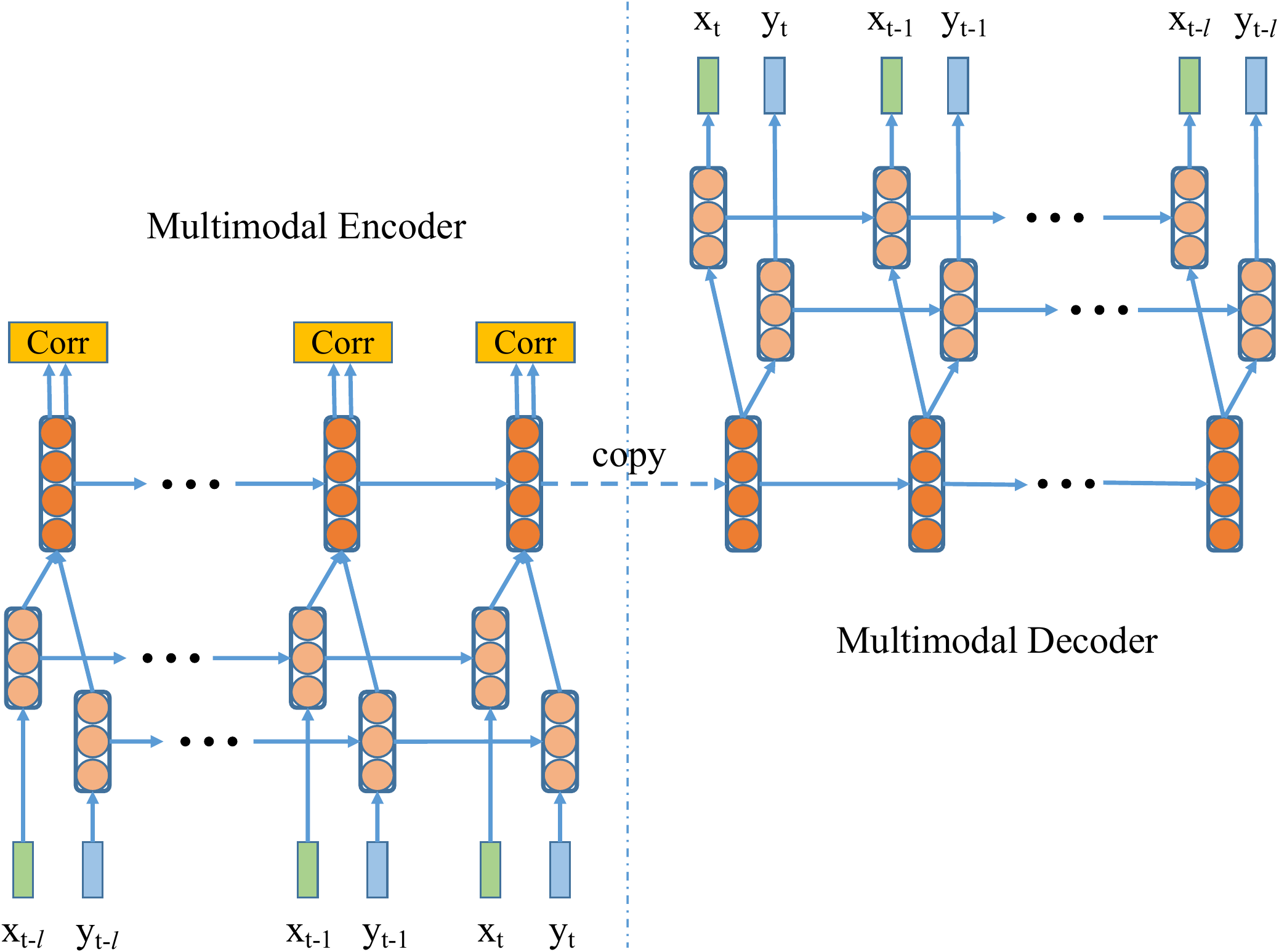}
\end{center}
\caption{Basic architecture of the proposed model}
\label{fig:fig2}
\vspace{-1em}
\end{figure}

We first describe the basic model architecture, as shown in Fig.~\ref{fig:fig2}. We implement an \textit{Encoder-Decoder} framework, which enables sequence-to-sequence learning \cite{sutskever2014sequence} and learning of sequence representations in an unsupervised fashion \cite{srivastava2015unsupervised}. Specifically, our model consists of two recurrent neural nets: the \textit{multimodal encoder} and the \textit{multimodal decoder}. The multimodal encoder is trained to map the two input sequences into a joint representation, \textit{i.e.}, a common space. The multimodal decoder attempts to reconstruct two input sequences from the joint representation obtained by the encoder. During the training process, the model learns a joint representation that retains as much information as possible from both modalities.

In our model, both the encoder and decoder are two-layer networks. The multimodal inputs are first mapped to separate hidden layers before being fed to a common layer called the \textit{fusion layer}. Similarly, the joint representation is first decoded to separate hidden layers before reconstruction of the multimodal inputs takes place.
 
\begin{figure*}
\begin{center}
\includegraphics[width=0.78\textwidth]{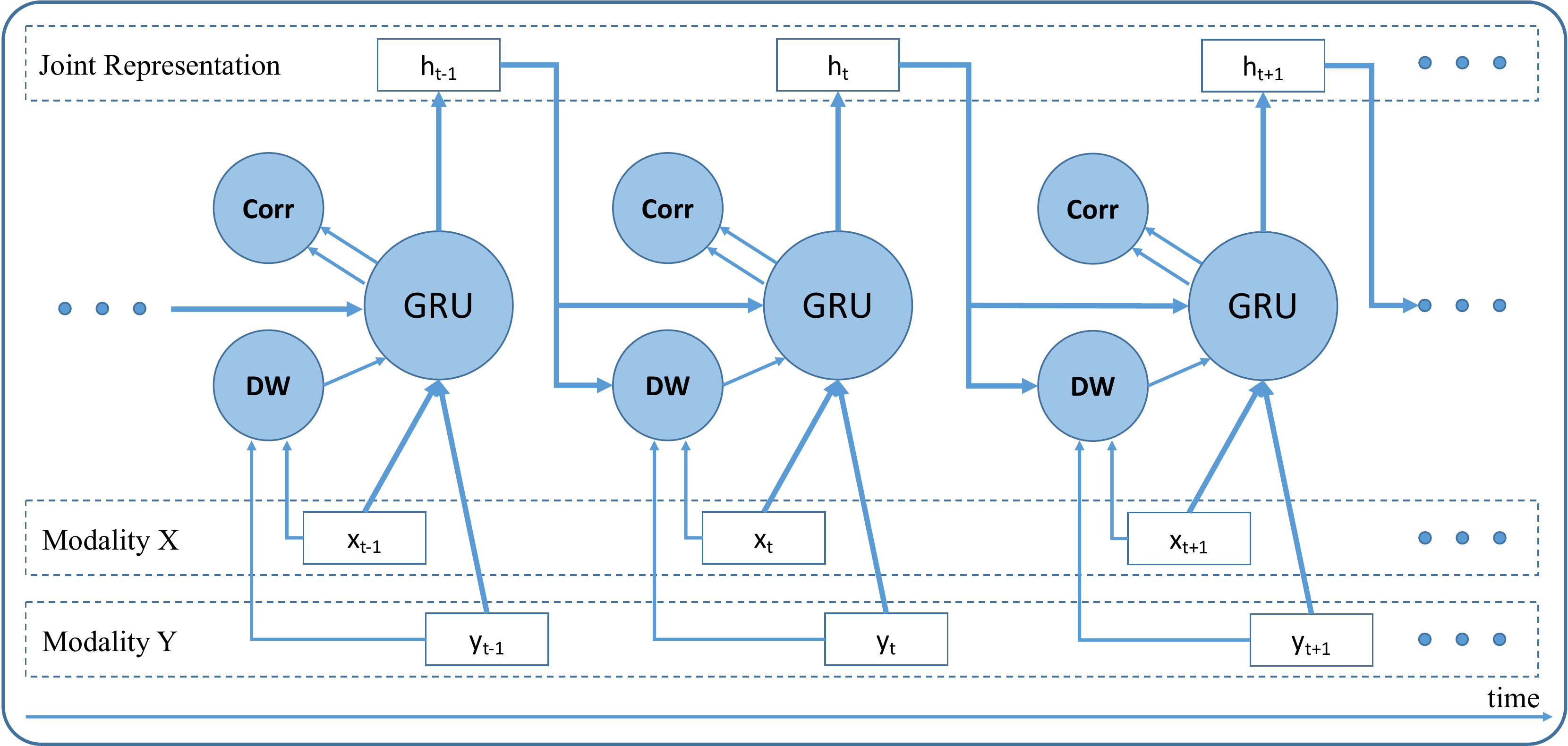}
\end{center}
\caption{The structure of the multimodal encoder. It includes three modules: Dynamic Weighting module (DW), GRU module (GRU) and Correlation module (Corr).}
\label{fig:encoder}
\vspace{-1em}
\end{figure*}

The standard Encoder-Decoder framework relies on the (reconstruction) loss function only in the decoder. As mentioned in Section \ref{sec:intro}, in order to obtain a better joint representation for temporal multimodal learning, we introduce two important components into the multimodal encoder, one that explicitly captures the correlation between the modalities, and another that performs dynamic weighting across modality representations. We also consider different types of reconstruction losses to enhance the capture of information within and between modalities.

Once the model is trained using a pair of multimodal inputs, the multimodal encoder plays the role of a feature extractor. Specifically, the activations of the fusion layer in the encoder at the last time step is output as the sequence feature representation. Two types of feature representation may be obtained depending on the model inputs: if both input modalities are present, we obtain their joint representation; on the other hand, if only one of the modalities is present, we obtain an ``enhanced" unimodal representation. The model may be extended to more than two modalities by maximizing the sum of correlations between all pairs of modalities. This can be implemented by adding more correlation modules to the multimodal encoder.

\subsection{Multimodal Encoder}
\label{sec:encoder}

The multimodal encoder is designed to fuse the input modality sequences into a common representation such that a coherent input is given greater importance, and the correlation between the inputs is maximized. Accordingly, three main modules are used by the multimodal encoder at each time step.

\begin{itemize}
	\item Dynamic Weighting module (DW): Dynamically assigns weights to the two modalities by evaluating the coherence of the incoming signal with recent past history. %by evaluating the agreement between the current input of each modality and the fused representation from the previous time step. %The module has two outputs, one used in the GRU module to assign weights to the two modalities, and the other used to control whether or not to compute the correlation-based loss.
    \item GRU module (GRU): Fuses the input modalities to generate the fused representation. The module also captures the temporal structure of the sequence using \textit{forget} and \textit{update} gates.   
	\item Correlation module (Corr): Takes the intermediate states generated by the GRU module as inputs to compute the correlation-based loss. %The module is triggered by 
%the DW module.
\end{itemize}

The structure of the multimodal encoder and the relationships among the three modules are illustrated in Fig. \ref{fig:encoder}. We now describe the implementation of these modules in detail.
% * <srig@acm.org> 2016-11-12T19:11:14.910Z:
%
% > We now describe the implementation of these modules in detail.
%
% To save space, we can avoid listing these modules first. 
%
% Xitong: my original idea is to make it more clear for readers to know the three main components of the encoder. We can skip that if we need to save space later.
% ^.

The \textbf{Dynamic Weighting module} assigns a weight to each modality input at a given time step according to an evaluation of its coherence over time. With reference to recent work on attention models \cite{BahdanauCB14}, our approach may be characterized as a soft attention mechanism that enables the model to focus on the modality with the more useful signal when, for example, the other is corrupted with noise. The dynamic weights assigned to the input modalities are based on the agreement between their current input and the fused data representation from the previous time step. This is based on the intuition that an input corrupted by noise would be less in agreement with the fused representation from the previous time step when compared with a ``clean" input. We use bilinear functions to evaluate the coherence scores $\alpha_t^1$ and $\alpha_t^2$ of the two modalities, namely:
% * <srig@acm.org> 2016-11-12T18:55:26.698Z:
%
% > the relevance between the current states of modalities and the previous fused states.
%
% Does this mean the model will pay less "attention" to key frames in video because they are different from previous frames ? This is not easy to justify.
%
% Xitong: Good point. Need to think about that...
% ^ <srig@acm.org> 2016-11-15T17:28:43.396Z.
$$\alpha_t^1 = x_t A_1 h_{t-1}^T,\ \ \  \alpha_t^2 = y_t A_2 h_{t-1}^T,$$
where $A_1 \in \mathbb{R}^{m\times d}, A_2 \in \mathbb{R}^{n\times d}$ are parameters learned during the training of the module. The weights of the two modalities is obtained by normalizing the scores using Laplace smoothing:
$$w_t^i = \frac{1 + \text{exp}(\alpha_t^i)}{2 + \sum_k \text{exp}(\alpha_t^k)}, \ \ i=1,2 $$

%The weights obtained above are also used as a trigger for the Correlation module. Given inputs $x_t$ and $y_t$, if any of the weights is smaller than a threshold $\eta$, we do not compute the correlation between these two modalities at time $t$. We empirically set $\eta = 0.1$ in our experiment.

\begin{figure}
\begin{center}
	\begin{subfigure}[Unimodal GRU]
    {\includegraphics[width=0.85\columnwidth]{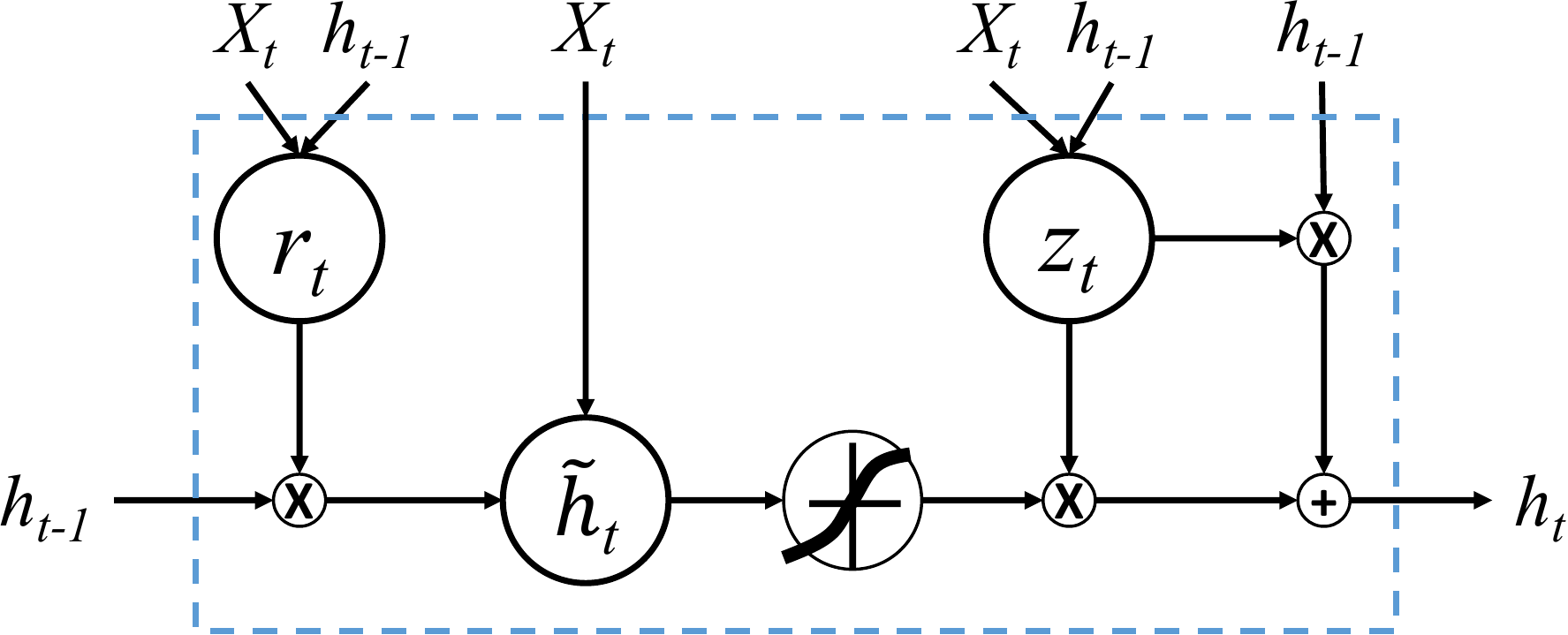}}
    \end{subfigure}
    \begin{subfigure}[Multimodal GRU]
    {\includegraphics[width=0.85\columnwidth]{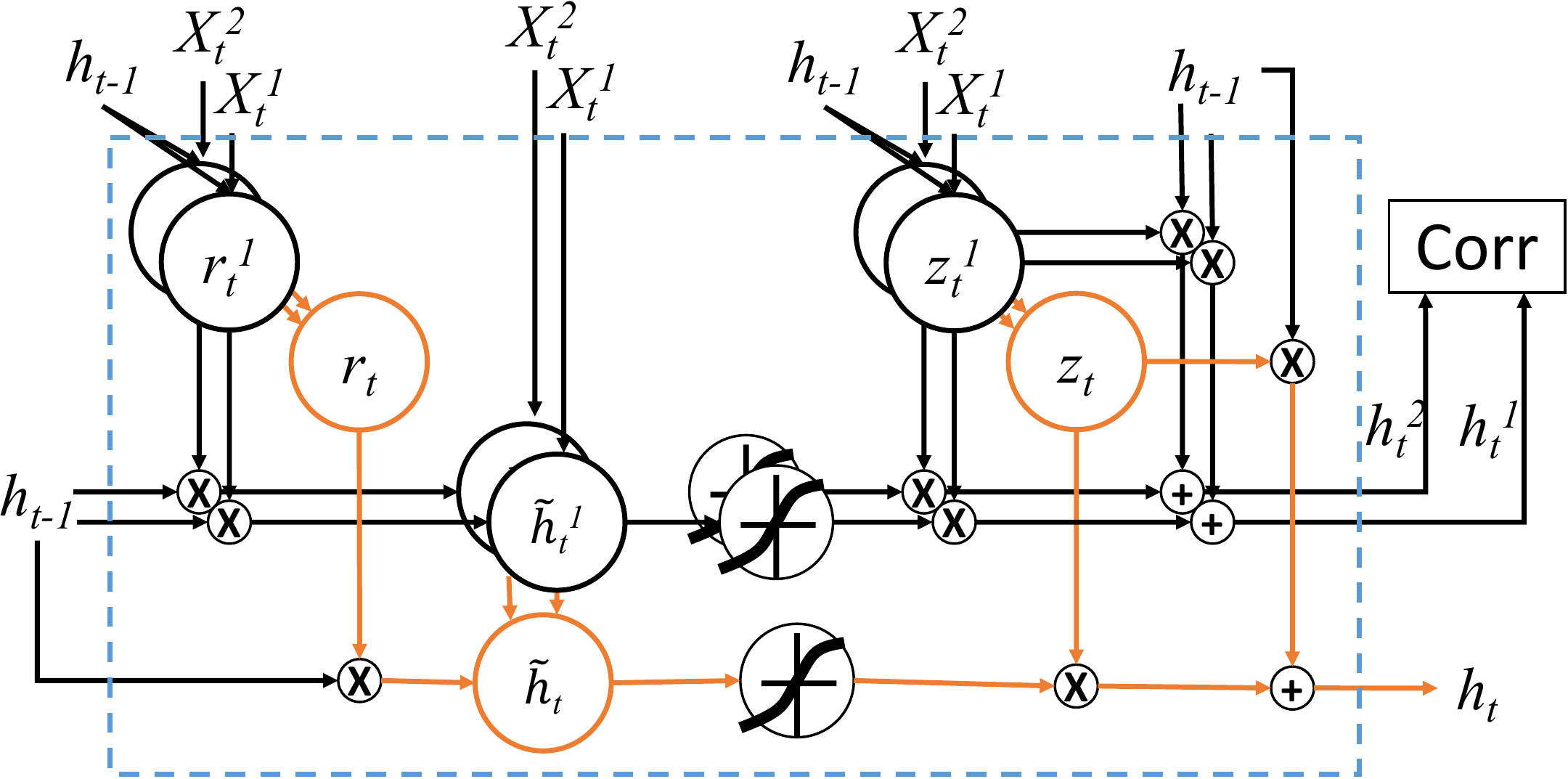}}
    \end{subfigure}
\end{center}
\caption{Block diagram illustrations of unimodal and multimodal GRU modules.}
\label{fig:GRU}
\end{figure}

The \textbf{GRU module} (see Fig. \ref{fig:GRU}(b)) is a multimodal extension of the standard GRU (see Fig. \ref{fig:GRU}(a)), and contains different gating units that modulate the flow of information inside the module. The GRU module takes $x_t$ and $y_t$ as input at time step $t$ and keeps track of three quantities, namely the fused representation $h_t$, and modality-specific representations $h_t^1$, $h_t^2$. The fused representation $h_t$ constitutes a single representation of historical multimodal input that propagates along the time axis to maintain a consistent concept and learn its temporal structure. The modality-specific representations $h_t^1$, $h_t^2$ may be thought of as projections of the modality inputs which are maintained so that a measure of their correlation can be computed. The computation within this module may be formally expressed as follows:
\begin{align}
r_t^i &=\sigma \left( \textbf{W}_r^i  X_t^i + \textbf{U}_r  h_{t-1} + b_r^i\right ),& i=1, 2 \\
z_t^i &=\sigma \left( \textbf{W}_z^i  X_t^i + \textbf{U}_z  h_{t-1} + b_z^i\right ),& i=1, 2 \\
\tilde{h}_t^i &= \varphi \left( \textbf{W}_h^i  X_t^i + \textbf{U}_h  (r_t^i \odot h_{t-1}) + b_h^i \right),&  i=1, 2
\end{align}
\begin{align}
r_t &=\sigma \left( \sum_{i=1}^2 w_t^i \left (\textbf{W}_r^i  X_t^i + b_r^i\right) + \textbf{U}_r  h_{t-1}) \right)\\
z_t &=\sigma \left( \sum_{i=1}^2 w_t^i \left (\textbf{W}_z^i  X_t^i + b_z^i\right) + \textbf{U}_z  h_{t-1}) \right)\\
\tilde{h}_t &= \varphi \left( \sum_{i=1}^2 w_t^i \left( \textbf{W}_h^i  X_t^i + b_h^i \right) + \textbf{U}_h  (r_t \odot h_{t-1}) \right)
\end{align}
\begin{align}
h_t^i &= (1 - z_t^i) \odot h_{t-1} +  z_t^i \odot \tilde{h}_t^i, & i=1,2\\
h_t &= (1 - z_t) \odot h_{t-1} + z_t \odot \tilde{h}_t
\end{align}

\noindent where $\sigma$ is the logistic sigmoid function and $\varphi$ is the hyperbolic tangent function, $r$ and $z$ are the input to the reset and update gates, and $h$ and $\tilde{h}$ represent the activation and candidate activation, respectively, of the standard GRU \cite{ChoMGBBSB14}.

Note that our model uses separate weights for the different inputs $X$ and $Y$, which differs from the approach proposed in \cite{ren2016look}. However, as we enforce an explicit correlation-based loss term in the fusing process, our model in principle can capture both the correlation across modalities, and specific aspects of each modality.

The \textbf{Correlation module} computes the correlation between the projections of the modality inputs  $h_t^1$and $h_t^2$ obtained from the GRU module. Formally, given $N$ mappings of two modalities $H_t^1 = \{h_{ti}^1\}_{i=1}^N$ and $H_t^2 = \{h_{ti}^2\}_{i=1}^N$ at time $t$, the correlation is calculated as follows:

$$corr(H_t^1, H_t^2) =
\frac{\sum_{i=1}^N (h_{ti}^1-\overline{H_t^1}) (h_{ti}^2-\overline{H_t^2})} {\sqrt{\sum_{i=1}^N (h_{ti}^1-\overline{H_t^1})^2 \sum_{i=1}^N (h_{ti}^2-\overline{H_t^2})^2}}
$$

\noindent where $\overline{H_t^1} = \frac{1}{N} \sum_i^N{h_{ti}^1}$ and $\overline{H_t^2} = \frac{1}{N} \sum_i^N{h_{ti}^2}$. We denote the correlation-based loss function as $L_{\text{corr}} = corr(H_t^1, H_t^2)$ and maximize the correlation between two modalities by maximizing this function. In practice, the empirical correlation is computed within a mini-batch of size $N$.

\subsection{Multimodal Decoder}
\label{sec:decoder}
The multimodal decoder attempts to reconstruct the individual modality input sequences $X$ and $Y$ simultaneously, from the joint representation $h_t$ computed by the multimodal encoder described above. By minimizing the reconstruction loss at training, the resulting joint representation retains as much information as possible from both modalities. In order to better share information across the modalities, we introduce two additional reconstruction loss terms into the multimodal decoder: \textit{cross-reconstruction} and \textit{self-reconstruction}. These two terms not only benefit the joint representation, but also improve the performance of the model in cases when only one of the modalities is present, as shown in Section \ref{sec:exp1}. In all, our multimodal decoder includes three reconstruction losses:

\begin{itemize}
\item \textbf{Fused-reconstruction loss}. The error in reconstructing $\tilde{x_i}$ and $\tilde{y_i}$ from joint representation $\tilde{h_i} = f(\tilde{x_i}, \tilde{y_i})$.
$$ L_{\text{fused}} = L(g(f(\tilde{x_i}, \tilde{y_i})), \tilde{x_i}) + \beta L(g(f(\tilde{x_i}, \tilde{y_i}), \tilde{y_i})$$

\item \textbf{Self-reconstruction loss}. The error in reconstructing $\tilde{x_i}$ from $\tilde{x_i}$, and $\tilde{y_i}$ from $\tilde{y_i}$.
$$ L_{\text{self}} = L(g(f(\tilde{x_i})), \tilde{x_i}) + \beta L(g(f(\tilde{y_i}), \tilde{y_i})$$

\item \textbf{Cross-reconstruction loss}. The error in reconstructing $\tilde{x_i}$ from $\tilde{y_i}$, and $\tilde{y_i}$ from $\tilde{x_i}$.
$$ L_{\text{cross}} = L(g(f(\tilde{y_i}), \tilde{x_i}) + \beta L(g(f(\tilde{x_i})), \tilde{y_i})$$
\end{itemize}

\noindent where $\beta$ is a hyperparameter used to balance the relative scale of the loss function values of the two input modalities, and $f, g$ denote the functional mappings implemented by the multimodal encoder and decoder, respectively. The objective function used to train our model may thus be expressed as:
$$\mathcal{L}=\sum_{i=1}^N \left( L_{\text{fused}}+L_{\text{cross}}+L_{\text{self}}\right )-\lambda L_{\text{corr}} $$where $\lambda$ is a hyperparameter used to scale the contribution of the correlation loss term, and $N$ is the mini-batch size used in the training stage. The objective function thus combines different forms of reconstruction losses computed by the decoder, with the correlation loss computed as part of the encoding process.  We use a stochastic gradient descent algorithm with an adaptive learning rate to optimize the objective function above.

\section{Empirical Analysis}
\label{sec:data}
In the following sections, we describe  experiments to demonstrate the effectiveness of CorrRNN at modeling temporal multimodal data. We demonstrate its general applicability to multimodal learning problems by evaluating it on multiple datasets, covering two different types of multimodal data (video-sensor and audio-video) and two different application tasks (activity classification and audio-visual speech recognition). We also evaluate our model in three \textit{multimodal learning settings} \cite{ngiam2011multimodal} for each task. We review these settings in Table \ref{tab:tab1}. 
\begin{table} [h!]
\begin{center}
\begin{tabular}{ |C{2.4cm} | C{1.3cm}| C{1.7cm}| C{1.3cm}|}
\hline
    & Feature Learning & Supervised Training & Testing \\ \hline
  Multimodal Fusion & $X+Y$ & $X+Y$ & $X+Y$ \\ \hline
  Cross Modality  & $X+Y$ & X & X \\
 Learning & $X+Y$ &Y & Y \\ \hline
  Shared Represe-  & $X+Y$ & X & Y \\ 
  ntation Learning & $X+Y$ & Y & X \\  
\hline \end{tabular}
\end{center}
\caption{Multimodal Learning settings, where $X$ and $Y$ are different input modalities}
\label{tab:tab1}
\end{table}

For each application task and dataset, the CorrRNN model is first trained in an unsupervised manner using both  the input modalities and the composite loss function described. The trained model is then used to extract %features $h_T (X+Y)$, $h_T^1 (X)$, and $h_T^2 (Y)$, 
the fused representation and the modality-specific representations of the data. Each of the multimodal learning settings is then implemented as a supervised classification task using a classifier, either an SVM or a logistic-regression classifier (in order to maintain consistency, the choice of classifier depends on the method involved in the benchmarking implemented). 

% In the multimodal fusion setting, the fused representation of the input modalities is used to train an SVM classifier. In the cross-modality and shared representation learning settings, the modality specific representations is used for training and testing the 

% Each of the multimodal learning settings is then implemented as a supervised classification task, wherein the trained model is used to extract features, and a simple classifier such as Logistic Regression or linear SVM is used for classification.

% We compare the following methods in all multimodal learning settings. The last two methods are both state-of-the-art models that combine the objective of reconstruction and correlation for multimodal learning.

% \begin{itemize}
% 	\item \textbf{Baseline}, no multimodal learning attempts, fused representation is obtained by concatenating input features,
%     \item \textbf{Linear CCA (CCA)}, canonical correlation analysis is performed on both modalities, fused representation is obtained by concatenation.
%     \item \textbf{Correlational Neural Network (CorrNet)}, the model proposed in \cite{chandar2015correlational},
%     \item \textbf{Deep Canonically Correlated Autoencoders (DCCAE)}, the model proposed in \cite{wang2015deep}.
% \end{itemize}

\subsection{Experiments on Video-Sensor Data}
\label{sec:exp1}
In this section, we apply the CorrRNN model to the task of human activity classification. For this purpose, we use the \textbf{ISI dataset} \cite{kumar2015fly}, a multimodal dataset in which 11 subjects perform seven actions related to an insulin self-injection activity. The dataset  includes egocentric video data acquired using a Google Glass wearable camera, and motion data acquired using an Invensense motion wrist sensor. Each subject's video and motion data is manually labeled and segmented into seven videos corresponding to the seven actions in the self-injection procedure. Each of these videos are further segmented into $ $ short video clips of fixed length. %\textbf{Due to the small size of the dataset, we extract overlapping video clips of fixed length from these videos, for a total of .... }

% \begin{itemize}
% \item 

% % \item \textbf{CMU-MMAC} \cite{spriggs2009temporal} contains multimodal data from human activity in a natural kitchen environment. As in \cite{spriggs2009temporal}, we use a subset of the dataset, corresponding to the first-person camera and motion sensor data for 7 subjects performing the ``Brownie'' activity. The available data is manually segmented and labeled with 29 action classes such as open fridge, beating eggs, stirring brownie mix, \textit{etc}.
% % \end {itemize}

\subsubsection{Implementation Details}
We first temporally synchronize the video and motion sensor data with the same sampling rate of 30 fps. We compute a 1024-dimensional CNN feature representation for each video frame using GoogLeNet \cite{szegedy2015going}. Raw motion sensor signals are smoothed by applying an averaging filter of width 4. Sensor features are obtained by computing the output of the last convolutional layer (layer 5) of a Deep Convolutional and LSTM (DCL) Network \cite{ordonez2016deep} pre-trained on the OPPORTUNITY dataset \cite{roggen2010collecting} to the smoothed sensor data input. The extracted features are a temporal sequence of $448$-dimensional elements. 

% For the CMU-MMAC dataset, following the approach used in \cite{spriggs2009temporal}, we perform PCA on both the CNN features and the smoothed sensor signals, and use the 100 and 45 most significant principal components, respectively, as features.

We build sequences from the video and sensor data, using a sliding window of 8 frames with a stride of $2$, sampled from a duration of $2$ seconds, resulting in $13,456$ sequences. 
These video and motion sequences are used to train the CorrRNN model, using stochastic gradient descent with the mini-batch size set to $256$. The values of $\beta$ and $\lambda$ were set to $1$ and $0.1$, respectively; these values were optimized using grid search methods. %, using the hyperparameter settings summarized in Table. \ref{tab:setting}.  
%Multiple training sequences are generated using a fixed stride.
% * <srig@acm.org> 2016-11-15T04:01:41.909Z:
%
% > Multiple training sequences are generated using a fixed stride.
%
% I added this generic sentence but in general we may need to provide some details about this.
%
% ^.
% The joint representation generated by CorrRNN is then used to train an SVM classifier for the activity recognition task.

% \begin{table} [h!]
% \begin{center}
% %\begin{tabular}{ |C{1.5cm} | {1.5cm} | C{1.8cm} | C{0.7cm} | C{0.7cm}|}
% \begin{tabular}{ |c | C{1.5cm} |c | c|c|}
% \hline
%     Dataset & Sequence Length & Hidden units  & $\beta$ & $\lambda$ \\
%     \hline
%     ISI & & 512 / 512 & 1 & 1 \\
%     \hline
%     CMU-MMAC & 15 & 128 / 256 & 10 & 0.1 \\
%     \hline 
% \end{tabular}
% \end{center}
% \caption{Hyperparameters settings for model training}
% \label{tab:setting}
% \end{table}

\subsubsection{Results}
\begin{figure}
\includegraphics[width=3in]{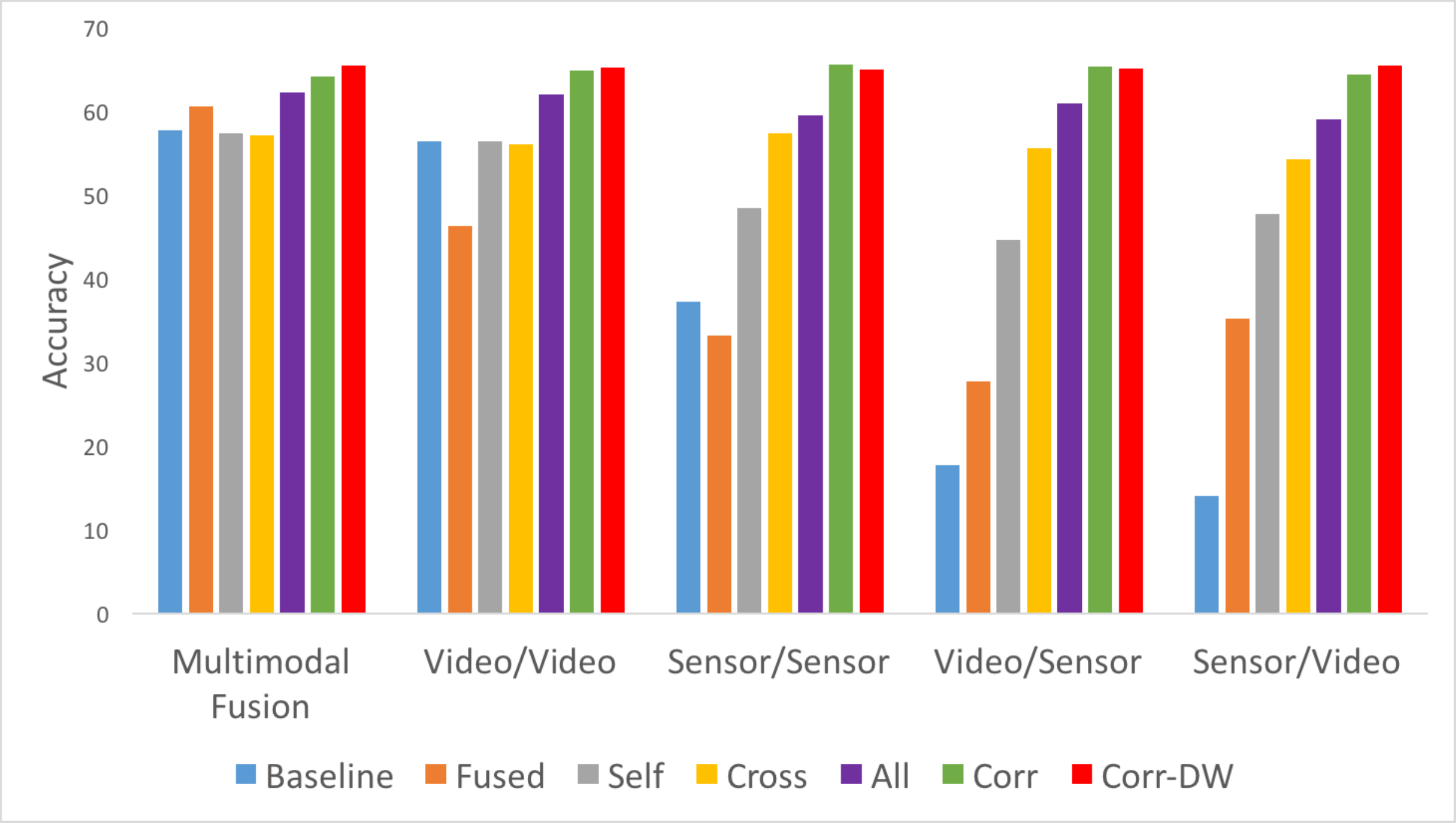}
\caption{Classification accuracy on the ISI dataset for different model configurations}
\label{fig:ISI}
\end{figure}

Figure~\ref{fig:ISI} shows the activity recognition accuracy of the proposed CorrRNN model.
We evaluate the contribution of each component in our model under the various multimodal learning settings listed in Table~\ref{tab:tab1}. In order to understand the contribution of different aspects of the CorrRNN design, we also evaluate different model configurations summarized in Table \ref{tab:config}.
The baseline results are obtained by first training a single layer GRU recurrent neural network with $512$ hidden units, separately for each modality. The $512$-dimensional hidden layer representations obtained from each network are then reduced to $256$ dimensions using PCA, and concatenated to obtain a $512$-dimensional fused representation. We observe that the fused representation obtained using CorrRNN significantly improves over this baseline fused representation.  

Each loss component contributes to better performance, especially in the settings of cross-modality learning and shared representation learning. Performance in the presence of poor fidelity or noisy modality (for instance, the motion sensor modality) is boosted by the inclusion of the other modality, due to the cross reconstruction loss component. Inclusion of the correlation loss and dynamic weighting further improves the accuracy. 

\begin{table} [t!]
\begin{center}
\begin{tabular}{ | l | l |}
\hline
    Config & Description  \\
    \hline
    Baseline & Single-layer GRU RNN per modality \\
    \hline
    Fused & Objective uses only  $L_{\text{fused}}$ term\\
    \hline 
        Self & Objective uses  $L_{\text{fused}}$ \& $L_{\text{self}}$ \\
    \hline 
        Cross & Objective uses   $L_{\text{fused}}$ \& $L_{\text{cross}}$ \\
    \hline 
        All & Objective uses  $L_{\text{fused}}$,$L_{\text{self}}$ \& $L_{\text{cross}}$ \\
    \hline 
        Corr & Objective uses all loss terms \\
    \hline 
        Corr-DW &  Objective uses all loss terms \& dyn. weights \\
    \hline 
\end{tabular}
\end{center}
\caption{CorrRNN model configurations evaluated}
\label{tab:config}
\end{table}

% The classification accuracies over three learning settings are plotted in Figure~\ref{fig:ISI}. 
% \begin{table}
% \begin{center}
% \begin{subfigure}[ISI]
% {\begin{tabular}{|c|c|c|c|c|c|}
% \hline
% & Multi- & \multicolumn{2}{c|}{Cross-modality} & \multicolumn{2}{c|}{Shared} \\
% & modal & \multicolumn{2}{c}{learning} & \multicolumn{2}{c}{representation} \\
% &fusion & & & \multicolumn{2}{c}{learning} \\
% & VS$\rightarrow$ VS & V $\rightarrow$ V & S $\rightarrow$S & V $\rightarrow$ S & S $\rightarrow$ V\\
% \hline
% Raw & & & & -&-\\
% \hline
% Fused & 83.17 & 78.83&45.33 & 32.83&57.83 \\
% \hline
% Self & 85.33& 84.33& 67.83& 60.50&74.80 \\
% \hline
% Cross & 89.00& 82.33& 88.50& 82.17& 85.00\\
% \hline
% Both & 86.33&88.00 &89.00 &87.00 & 88.00\\
% \hline
% Corr & 86.33&85.00 &88.00 &85.00 &85.67 \\
% \hline
% Attention & & & & & \\
% \hline
% \end{tabular}}
% \end{subfigure}
% \begin{subfigure}[CMU-MMAC]
% {\begin{tabular}{|c|c|c|c|c|c|}
% \hline
% & VS$\rightarrow$ VS & V $\rightarrow$ V & S $\rightarrow$S & V $\rightarrow$ S & S $\rightarrow$ V\\
% \hline
% Raw & 49.23 & 42.77 & 33.51 & 3.84 & 11.05\\
% \hline
% Fused & 52.57 & 47.13 & 32.39 & 3.5 & 20.53\\
% \hline
% Self & 52.70 & 47.67 & 34.79 & 18.95 & 3.67 \\
% \hline
% Cross & 43.23 & 43.26 & 27.08 & 15.89 & 25.09 \\
% \hline
% Both & & & & & \\
% \hline
% Corr & & & & & \\
% \hline
% Attention & & & & & \\
% \hline
% \end{tabular}}
% \end{subfigure}
% \end{center}
% \caption{Classification performance of the proposed *** for task classification on (a) ISI and (b) CMU-MMAC datasets.}
% \end{table}

%\paragraph*{ Impact of Correlation Loss} 
In Table \ref{tab:corr}, we compare the correlation between the projections of the modality inputs for different model configurations. This measure of correlation is computed as the mean encoder loss over the training data in the final training epoch, divided by the number of hidden units in the fusion layer. These values demonstrate that the use of the correlation-based loss term maximizes the correlation between the two projections, leading to a richer joint and shared representations. 
\begin{table}
\begin{center}
\begin{tabular} {|l|c|}
	\hline
    Configuration & Correlation \\ 
    \hline
    Fused & 0.46 \\
    Self &  0.67\\
    Cross &  0.76\\
    \hline
    Corr &  0.95\\
    Corr-DW & 0.93 \\
    \hline
\end{tabular}
\end{center}
\caption{Normalized correlation for different model configurations}
\label{tab:corr}
\end{table}

% \paragraph*{Dynamic Weighting}

%\paragraph*{Missing Value?}

%\paragraph*{Long-term dependencies}

% We also qualitatively investigate the aforementioned features by embedding the learned features in 2D using \textit{t}-SNE \cite{van2008visualizing}. The resulting visualizations are plotted in Figure. Overall, even though we 

\subsection{Experiments on Audio-Video Data}
\label{sec:exp2}
The task of audio-visual speech classification using multimodal deep learning has been well studied in the literature~\cite{hu2016temporal,ngiam2011multimodal}. In this section, we focus on comparing the performance of the proposed model with other published methods on the AVLetters and CUAVE datasets:
\begin{itemize}
\item \textbf{AVLetters} \cite{matthews2002extraction} includes audio and video of 10 speakers uttering the English alphabet three times each. We use the videos corresponding to the first two times for training ($520$ videos) and the third time for testing ($260$ videos). This dataset provides pre-extracted lip regions scaled to $60 \times 80$ pixels for each video frame and $26$-dimensional \textit{Mel-Frequency Cepstrum Coefficient} (MFCC) features for the audio.

\item \textbf{CUAVE} \cite{patterson2002cuave} consists of videos of $36$ speakers pronouncing the digits 0-9. Following the protocol in~\cite{ngiam2011multimodal}, we use the first part of each video, containing the frontal facing speakers pronouncing each digit $5$ times. The even-numbered speakers are used for training, and the odd-numbered speakers are used for testing. The training dataset contains $890$ videos and the test data contains $899$ videos. We pre-processed the video frames to extract only the region of interest containing the mouth, and rescaled each image to $60 \times 60$ pixels. The audio is represented using $26$-dimensional MFCC features.

\end{itemize}
\subsubsection{Implementation Details}
We reduced the dimensionality of the video features of both the datasets to $100$ using PCA whitening, and concatenated the features representing every $3$ consecutive audio samples, in order to align the audio and the video data. 
% * <srig@acm.org> 2016-11-15T18:31:18.447Z:
%
% > and concatenated the features representing every $3$ consecutive audio samples
%
% Do we need to mention that this was done to obtain some parity in feature dimensionality across modes ?  Also I think we should mention how we bring about synchronization between modes.
%
% ^ <radha.cr@gmail.com> 2016-11-15T20:24:04.502Z:
%
% For both the datasets the number of audio samples was approximately 3 times the number of video frames. So when we concatenate it not only increases the dimensionality but also aligns the audio and video,
%
% ^.
In order to train the CorrRNN model, we generated sequences with length 8 using a stride of 2. Training was performed using stochastic gradient descent with the size of the mini-batch set to 32. The number of hidden units in the hidden layers was set to 512. %The value of $\beta$ is set to $1$ and the value of $\lambda$ to $3$. 
After training the model in an unsupervised manner, the joint representation generated by CorrRNN is treated as the fused feature. Similar to \cite{hu2016temporal}, we first break down the fused features of each speaking example into one and three equal slices and perform mean-pooling over each slice. The  mean-pooled features for each slice are then concatenated and used to train a linear SVM classifier in a supervised manner.

\subsubsection{Results}
Table~\ref{tbl:audio-visual1} showcases the classification performance of the proposed CorrRNN model using the Corr-DW configuration on the AVLetters and the CUAVE datasets. The fused representation of the audio-video data generated using the CorrRNN model is used to train and test an SVM classifier. We observe that the CorrRNN representation leads to more accurate classification than the representation generated by non-temporal models such as Multimodal deep autoencoder (MDAE), multimodal deep belief networks (MDBN), and the multimodal deep Boltzmann machines (MDBM). This is because the CorrRNN model is able to learn the temporal dependencies between the two modalities. CorrRNN also outperforms conditional RBM (CRBM), and RTMRBM models due to the incorporation of the correlational loss and the dynamic weighting mechanism. 

The CorrRNN model also produces rich representations for each modality, as demonstrated in the cross-modality and shared representation learning experimental results in Table~\ref{tbl:audio-visual2}. Indeed, there is a significant improvement in accuracy from using CorrRNN features relative to the scenarios where only the raw features for both audio and video modalities are used, and this improvement holds for both the datasets. For instance, the accuracy improves by more than two times on the CUAVE dataset by learning the video features with both audio and video, compared to learning only with the video features. In the shared representation learning experiments, we learn the feature representation using both the audio and video modalities, but the supervised training and testing are performed using different modalities. The results show that the CorrRNN model captures the correlation between the modalities very well.   

In order to evaluate the robustness of the CorrRNN model to noise, we added white Gaussian noise at 0dB SNR to the original audio signal in the CUAVE dataset. Unlike prior models whose performance degrades significantly ($12-20\%$) due to presence of noise , there is only a minor decrease of about $5\%$ in the accuracy of the CorrRNN model, as shown in Table~\ref{tbl:audio-visual3}. This may be ascribed to the richness of the cross-modal information embedded in the fused representation learned by CorrRNN. 

\begin{table}
\begin{center}
\begin{tabular}{|c||c|c|}
\hline
 Method & \multicolumn{2}{c|}{Accuracy} \\
 \cline{2-3}
       & AVLetters & CUAVE \\
\hline
 %Concatenated & $68.08$ & $89.21$\\
 %features     &         &        \\
\hline
MDAE~\cite{ngiam2011multimodal} & $62.04$ &  $66.70$ \\
\hline
MDBN~\cite{srivastava2012multimodal} &$63.2$ &$67.20$ \\
\hline
MDBM~\cite{srivastava2012multimodal} & $64.7$&$69.00$\\
\hline
RTMRBM~\cite{hu2016temporal} & $66.04$ & -\\
\hline
CRBM~\cite{amer2014multimodal} & $67.10$ & $69.10$ \\
\hline
\textbf{CorrRNN} & $\mathbf{83.40}$ & $\mathbf{95.9}$\\ 
\hline
\end{tabular}
\end{center}
\caption{Classification performance for audio-visual speech recognition on the AVLetters and CUAVE datasets, compared to the best published results in literature, using the fused representation of the two modalities.}
\label{tbl:audio-visual1}
\end{table}

\begin{table}
\begin{center}
\begin{tabular}{|c||c||c||c||c|c|}
\hline
 & Train & Method &\multicolumn{2}{c|}{Accuracy} \\
\cline{4-5}
       &   /Test    & & AVLetters & CUAVE \\
\hline
Cross- &Video & Raw& 38.08& 42.05\\
\cline{3-5}
modality&/Video & CorrRNN& 81.85& 96.22\\
\cline{2-5}
learning & Audio& Raw&57.31 & 88.32\\
\cline{3-5}
& /Audio&CorrRNN &85.33 & 96.11\\
\hline
\hline
Shared& Video& MDAE& -&24.30\\
\cline{3-5}
represe-& /Audio&CorrRNN &85.33 &96.77\\
\cline{2-5}
ntation & Audio& MDAE&- &30.70 \\
\cline{3-5}
learning& /Video&CorrRNN &81.85 & 96.33\\
\hline
\end{tabular}
\end{center}
\caption{Classification accuracy for the cross-modality and shared representation learning settings. MDAE results from~\cite{ngiam2011multimodal}.}
\label{tbl:audio-visual2}
\end{table}

\begin{table}
\begin{center}

\begin{tabular}{|c|c|c|}
\hline
Method & \multicolumn{2}{c|}{Accuracy } \\
\cline{2-3}
& Clean Audio & Noisy Audio \\
\hline
MDAE & 94.4 & 77.3 \\
\hline
Audio RBM & 95.8& 75.8\\
\hline
MDAE + Audio RBM & 94.4 & 82.2 \\ 
\hline
CorrRNN & \textbf{96.11} & \textbf{90.88} \\
\hline
\end{tabular}
\end{center}
\caption{Classification accuracy for audio-visual speech recognition on the CUAVE dataset, under clean and noisy audio conditions. White Gaussian noise is added to the audio signal at 0dB SNR. Baseline results from~\cite{ngiam2011multimodal}.}
\label{tbl:audio-visual3}
\end{table}
% Long-term dependencies
% Supervised finetuning

\section{Conclusions}
\label{sec:conclude}

In this paper, we have proposed CorrRNN, a new model for multimodal fusion of temporal inputs such as audio, video and sensor data. The model, based on an Encoder-Decoder framework, learns joint representations of the multimodal input by exploiting correlations across modalities. The model is trained in an unsupervised manner (\textit{i.e.}, by minimizing an input-output reconstruction loss term and maximizing a cross-modality-based correlation term) which obviates the need for labeled data, and incorporates GRUs to capture long-term dependencies and temporal structure in the input. We also introduced a dynamic weighting mechanism that allows the encoder to dynamically modify the contribution of each modality to the feature representation being computed.  We have demonstrated that the CorrRNN model achieves state-of-the-art accuracy in a variety of temporal fusion applications. In the future, we plan to apply the model to a wider variety of multimodal learning scenarios. We also plan to extend the model to seamlessly ingest asynchronous inputs.

{\small
\bibliographystyle{ieee}
\bibliography{TemporalFusion}
}

\end{document}